\documentclass[runningheads]{llncs}
\usepackage[T1]{fontenc}
\usepackage{graphicx,verbatim}

\usepackage{booktabs}
\usepackage{multirow}

\usepackage{amsfonts}
\usepackage{amssymb}
\usepackage{amsmath}
\usepackage{subfig}
\usepackage{mathtools}
\usepackage{hyperref}
\usepackage{pifont}
\usepackage[table]{xcolor}

\begin{document}
\title{Radiomic fingerprints for knee MR images assessment}
\titlerunning{Radiomic fingerprints for knee MR images assessment}

%

\author{Yaxi Chen\inst{1,2}\orcidID{0009-0007-5906-899X} 
\and Simin Ni\inst{3}\orcidID{0009-0007-2780-6118} 
\and Shaheer U. Saeed\inst{2,4}\orcidID{0000-0002-5004-0663} 
\and Aleksandra Ivanova\inst{3}\orcidID{0009-0000-4113-8928} 
\and Rikin Hargunani\inst{5}\orcidID{0000-0002-0953-8443} 
\and Jie Huang\inst{1}\orcidID{0000-0001-7951-2217} 
\and Chaozong Liu\inst{3,5}\orcidID{0000-0002-9854-4043} 
\and Yipeng Hu\inst{2,4}\orcidID{0000-0003-4902-0486}
}

\authorrunning{Y. Chen et al.}

%

\institute{Mechanical Engineering Department, University College London, London, UK \and Hawkes Institute, University College London, London, UK \and Institute of Orthopaedic \& Musculoskeletal Science, University College London, Royal National Orthopaedic Hospital, Stanmore, UK \and Department of Medical Physics and Biomedical Engineering, University College London, London, UK \and Royal National Orthopaedic Hospital, Stanmore, UK}


\maketitle              
\begin{abstract}
Accurate interpretation of knee MRI scans relies on expert clinical judgment, often with high variability and limited scalability. Existing radiomic approaches use a fixed set of radiomic features (the ''signature''), selected at the population level and applied uniformly to all patients. While interpretable, these signatures are often too constrained to represent individual pathological variations. As a result, conventional radiomic-based approaches are found to be limited in performance, compared with recent end-to-end deep learning alternatives without using interpretable radiomic features. 

We argue that the individual-agnostic nature in current radiomic selection is not central to its intepretability, but is largely responsible for the poor generalization in our application. Here, we propose a novel radiomic fingerprint framework, in which a radiomic feature set (the ''fingerprint'') is dynamically constructed for each patient, selected by a deep learning model. Unlike the existing radiomic signatures, our fingerprints are derived on a per-patient basis by predicting the feature relevance in a large radiomic feature pool, and selecting only those that are predictive of clinical conditions for individual patients. The radiomic-selecting model is trained simultaneously with a low-dimensional (considered relatively explainable) logistic regression for downstream classification. 

We validate our methods across multiple diagnostic tasks including general knee abnormalities, anterior cruciate ligament (ACL) tears, and meniscus tears, demonstrating comparable or superior diagnostic accuracy relative to state-of-the-art end-to-end deep learning models. Perhaps more importantly, we show that the interpretability inherent in our approach facilitates meaningful clinical insights and potential biomarker discovery, with detailed discussion, quantitative and qualitative analysis of real-world clinical cases to evidence these advantages. The codes are available at: \url{https://github.com/YaxiiC/RadiomicsFingerprint.git}

\end{abstract}
\section{Introduction}

Common and clinically significant knee injuries, such as anterior cruciate ligament (ACL) tears and meniscal tears, are routinely evaluated using MRI~\cite{salzler2015state} as gold standard due to its exceptional capability to visualize detailed anatomical structures and pathological changes non-invasively. MRI provides critical insights into the precise location, depth, and patterns of tissue injuries, making it indispensable in clinical decision-making and treatment planning. Beyond detection, MRI enables detailed assessment of injury characteristics, including the distinction between partial and complete ACL tears, as well as the classification of meniscal damage, which guides decisions between repair and resection~\cite{al2024magnetic}. These insights are critical for selecting between conservative management, such as physiotherapy, and surgical options, including ACL reconstruction or meniscal repair and meniscectomy. With rising imaging demand, automated analysis tools are increasingly needed to support timely and informed clinical decision-making.

Existing approaches for automated knee MRI assessment generally fall into two broad paradigms: classical radiomics and deep learning (DL). Radiomics involves the extraction of hand-crafted quantitative features from regions of interest (ROI) in the images, capturing characteristics such as intensity distributions, texture patterns, shape descriptors, and wavelet coefficients~\cite{zhang2023radiomics}. Such radiomic models are valued for their interpretability – each feature has a well-defined meaning (intensity, roughness, size, etc.) that can be understood by clinicians. A radiologist can potentially relate a feature’s value to an aspect of the anatomy or pathology, making the decision process more transparent. The trade-off, however, is that radiomics use a fixed, hand-crafted feature set that may not capture the full complexity of imaging patterns, and these simpler models often yield lower diagnostic accuracy than state-of-the-art DL in practice~\cite{ardakani2022interpretation,lavrova2024review}. Additionally, radiomic features can be sensitive to imaging protocol variations, raising generalizability concerns if not carefully standardized.

In contrast, DL methods learn their own task-specific features directly from images, typically achieving higher accuracy in various downstream tasks. Most previous studies in this area have utilized end-to-end convolutional neural networks (CNNs). For instance, Bien et al.~\cite{bien2018deep} utilized an end-to-end classifier and Tsai et al.~\cite{tsai2020knee} applied an a lightweight efficiently layered network (ELNet). These DL models demonstrated high diagnostic performance for knee pathology detection, often approaching expert-level sensitivity and specificity. The decision mechanism of a CNN is distributed across many abstract features, making it difficult to interpret which image characteristics led to a given prediction. While methods like class activation mappings (CAM)~\cite{bien2018deep} have been used to enhance interpretability, a major limitation of these DL models remains their “black box” nature. The models generally lack the transparent feature-by-feature reasoning that radiomics offers. The lack of transparency and thus human understanding of how image-level features are learned and decisions are made remains an interesting challenge. Numerous studies in explainable AI emphasize this concern and call for transparent models in high-stakes domains~\cite{chen2022explainable}. 

Recent studies have suggested that integrating hand-crafted features with deep features or using hybrid models can yield more robust results in medical imaging tasks~\cite{kobayashi2021observing}. This gap - between transparent but rigid radiomics and accurate but inscrutable DL - has motivated our proposed patient-specific radiomic fingerprint framework. Instead of relying on a static, one-size-fits-all feature set, our method dynamically tailors the radiomic features to each individual case via a neural network-based feature selection mechanism. Specifically, we employ a feature-weighting neural network that analyzes the patient’s MRI and assigns an importance weight (or selection probability) to each feature in a large pool of candidate radiomic features. In this way, the model learns to select an optimal subset of features for each patient – effectively constructing a personalized radiomics fingerprint on the fly, tuned to the particular imaging characteristics of that patient’s knee. These selected features then serve as inputs to a downstream logistic regression model, which produces the final classification (e.g., presence or absence of an ACL tear). The feature-weighting network and the logistic regression classifier are trained jointly in an end-to-end fashion, under supervision of the known condition labels. This integration ensures that the network learns to choose feature subsets that maximize the downstream diagnostic performance, while the logistic regression provides a simple, interpretable mapping from those features to the prediction. This study builds on the previous work~\cite{chen2025patient}, which integrated radiomics with reconstructed patient-specific healthy baselines and emphasized visual interpretability. In contrast, this work focuses on maximising diagnostic performance while exploring interpretability based solely on radiomic fingerprints. We expand the feature pool to include both first- and higher-order texture features to build a comprehensive radiomic descriptors, enhancing fingerprint expressiveness and diagnostic utility.

Our approach achieves the dual goals of preserving interpretability and improving clinical performance in automated knee MRI diagnosis. The DL–guided dynamic feature selection empowers the model with patient-specific flexibility and enhanced accuracy. The result is a tailored radiomic fingerprint for each patient that is both explainable and predictive. We summarize our key contributions as follows: 1) A patient-specific radiomic fingerprint approach that significantly improves interpretability and performance; 2) Validation demonstrating comparable or superior diagnostic accuracy compared to traditional DL methods in multiple knee MRI classification tasks; and 3) An open-source implementation enabling reproducibility and future advancements in interpretable medical imaging.

\section{Methods}
\label{sec:method}

Our method is built around the concept of radiomic fingerprint - a subject-specific, dynamically constructed representation derived from a large pool of radiomic features. 
The radiomic fingerprint in our framework is computed by predicting a dense relevance vector over a comprehensive set of hand-crafted radiomic features extracted from multi-scale, multi-view subregions of the image. These feature weight are modeled as continuous-valued probabilities, which are thresholded at inference time to yield a binary selection for individual radiomic features. The final fingerprint is therefore a sparse, interpretable, yet highly discriminative set of radiomic features tailored to each patient. Because selection occurs over a rich feature pool, the effective feature space of the fingerprint is significantly larger than that of any fixed radiomics signature, supporting greater expressivity and finer-grained diagnostic encoding.

The following subsections describe the feature extraction process, the neural network architecture used for relevance prediction, the construction of the fingerprint vector, and the downstream classification procedure.

\paragraph{\textbf{Feature Extraction:}}  
Given an image ROI $\mathbf{x} \in \mathbb{R}^{H \times W \times D}$, where $H$, $W$, and $D$ denote its spatial dimensions. To represent the spatially local variation, the ROI is further divided into $J$ non-overlapping patches. From each patch, a collection of 3D radiomic features are computed independently on three different image volumes: axial, sagittal, and coronal views. Further details on the definitions of radiomic features, ROIs, and registration aligning them from the three sequences are described in Sec.~\ref{subsec:dataset}.

Let \( f_{j,k}^{(v)} \) denote the \( k^{th} \) feature extracted from the \( j^{th} \) patch in \( v^{th} \) image sequence, where \( k = 1, \dots, K \), \( j = 1, \dots, J \) and $v\in\{1,2,3\}$. Thus, the feature vector $\mathbf{f}_j$ for the $j^{th}$ patch:
\[
\mathbf{f}_j = [f_{j,1}^{(1)},...,f_{j,K}^{(1)}, f_{j,1}^{(2)},...,f_{j,K}^{(2)}, f_{j,1}^{(3)},...,f_{j,K}^{(3)}]^\top \in \mathbb{R}^{3K}
\]
Now, the full feature vector $\mathbf{f}$ assembles $\mathbf{f}_j$ vectors from all $J$ patches:
\[
\mathbf{f} = [\mathbf{f}_1^\top, \mathbf{f}_2^\top, ..., \mathbf{f}_J^\top]^\top \in \mathbb{R}^{3JK}
\]
For notational brievity, the full feature vector $\mathbf{f}$ is henceforth denoted with features $f_i$ using a single subscript $i = 1, ..., 3JK$:
\[
\mathbf{f} = [f_1, f_2, ..., f_{3JK}]^\top
\]

\paragraph{\textbf{Dynamic Feature Selection via Neural Network:}}
A neural network $G_{\omega}(\mathbf{x})$, parameterized by $\omega$, takes the entire ROI $x$ as input and predicts a relevance score vector $\mathbf{q}$ for all computed radiomic features $\mathbf{f}$:

\[
\mathbf{q} = G_{\omega}(\mathbf{x}) = [q_1, q_2, ..., q_{3JK}]^\top \in \mathbb{R}^{3JK}
\]
Each scalar weight \( q_i \in [0,1] \) quantifies the importance of individual features \( f_i \), with respect to the downstream task.
During training, the relevance scores $\mathbf{q}$ are applied to the corresponding features via element-wise multiplication:

\[
\mathbf{f}^{\text{w}} = \mathbf{f} \odot \mathbf{q} = [f_1 \cdot q_1, f_2 \cdot q_2, \dots, f_{3JK} \cdot q_{3JK}]^T.
\]

\paragraph{\textbf{Downstream Classification:}}  
The weighted feature vector $\mathbf{f}^{\text{w}}$ represents the soft assignment of individual features. Normalisation is not applied to allow flexibility in determining empirical threshold values (Sec.~\ref{subsec:dataset}), when taking into account inference-time application-specific requirement, e.g. size of selected feature set. This vector is passed directly to a clinical task classifier $R_{\psi}$, parameterized by $\psi$, to predict the final diagnostic class probabilities $\hat{\mathbf{c}}$:

\[
\hat{\mathbf{c}} = R_{\psi}(\mathbf{f}^{\text{w}}).
\]
This classifier is designed as a low-dimensional function without higher-degree terms, for its intepretability, such as the logistic regression used in this study, and is commonly adopted in classical radiomic signature algorithms for the same considerations.

\begin{figure}
    \centering
     \includegraphics[width=1\textwidth]{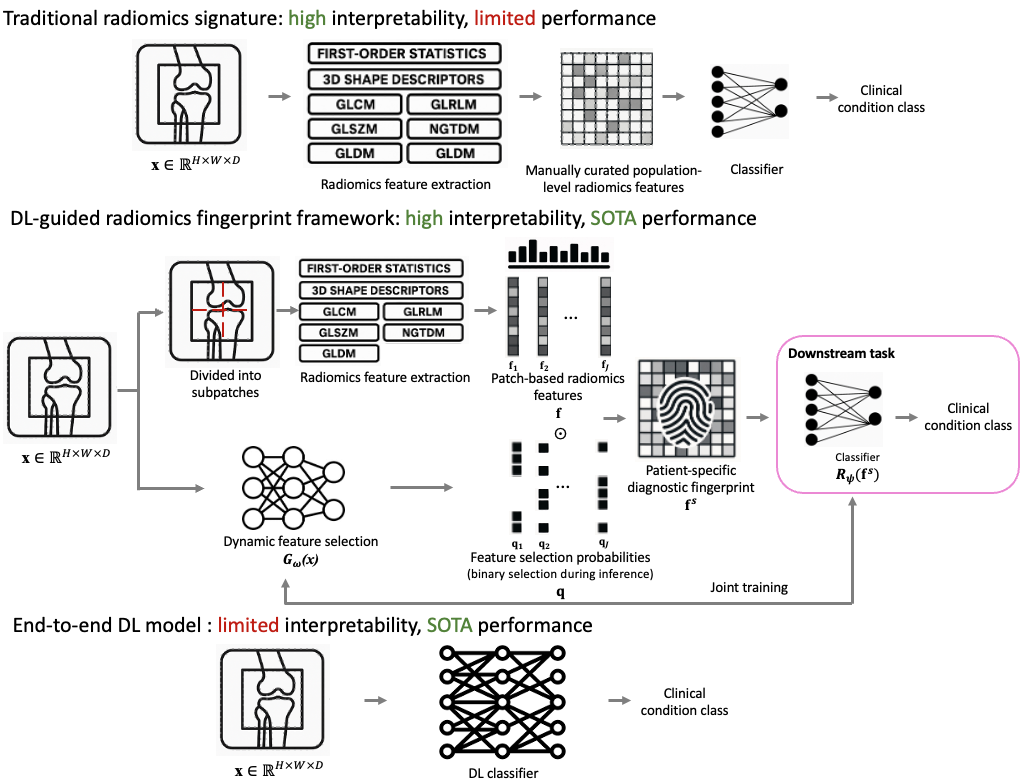}
    \caption{Comparison between the traditional radiomic signature approaches, our proposed radiomic fingerprint framework, and example end-to-end DL models.}
    \label{fig1}
\end{figure}

\paragraph{\textbf{Training:}}  
To summarise, features $\mathbf{f}$ are extracted from all patches across all three view-differing MR sequences. The feature-weighting network $G_\omega$ takes the full ROI as input and predicts the corresponding weight $\mathbf{q}$. The weighted feature vector $\mathbf{f}^{\text{w}} = \mathbf{f} \odot \mathbf{q}$ is the input of the classifier $R_\psi$. The parameters of both the network and the classifier are optimized jointly by minimizing a cross-entropy loss $\mathcal{L}(\hat{\mathbf{c}}, \tilde{\mathbf{c}})$ over the training set, with available class labels $\tilde{\mathbf{c}}$:

\[
(\omega^*, \psi^*) = \arg\min_{\omega, \psi} \sum \mathcal{L}(R_\psi(\mathbf{f} \odot G_{\omega}(\mathbf{x})), \tilde{\mathbf{c}}).
\]







\paragraph{\textbf{Feature Selection at Inference:}} 
At inference time, the continuous feature weights predicted by the feature-weighting network $G_\omega$ are thresholded to obtain a binary feature selection mask, $q_i^{(b)} = 1$ if $q_i \geq T$, 0 otherwise, where \( T \) is a predefined threshold. The resulting patient-specific fingerprint $\mathbf{f}^s$ is therefore obtained: $\mathbf{f}^s = \mathbf{f} \odot \mathbf{q}^{(b)}$, where 
$\mathbf{q}^{(b)} = [q_1^{(b)}, q_2^{(b)}, ..., q_{3JK}^{(b)}]^\top $.
This sparse fingerprint is then used for final prediction on new test data:
\[
\hat{\mathbf{c}} = R_{\psi}(\mathbf{f}^{\text{s}}).
\]

\section{Experiments and Results}

\subsection{Dataset and Implementation Details}

\label{subsec:dataset}
\noindent\textbf{Dataset and Preprocessing:}  
Our methodology is demonstrated on the MRNet dataset~\cite{bien2018deep}, comprising 1,370 3D knee MRI scans collected across axial, coronal, and sagittal planes. Each exam is annotated for three diagnostic labels: general abnormalities, ACL tear, and meniscal tear. We use 1,130 studies for development set and 120 for test. All images are preprocessed by resizing them to \(32 \times 128 \times 128\) voxels. To ensure consistent ROI definition and patch division, we apply affine registration using NiftyReg~\cite{modat2010fast}, and evaluate performance with and without this step. The ROI is defined as a subvolume encompassing \(50\%\)  of the original image depth, \(30\%\) of the height, and \(50\%\) of the width, which is empirically determined by radiologists to effectively cover anatomical regions commonly associated with ACL and meniscus pathology.

\noindent\textbf{Feature Extraction and dynamic selection:}  Radiomic features are extracted using \texttt{PyTorchRadiomics}, a PyTorch-based implementation built upon the PyRadiomics library~\cite{van2017computational}. We compute a comprehensive set of features, including \textit{First-Order Statistics} (19 features), \textit{3D Shape Descriptors} (16 features), and texture-based features such as \textit{Gray Level Co-occurrence Matrix (GLCM)} (24 features), \textit{Gray Level Run Length Matrix (GLRLM)} (16 features), \textit{Gray Level Size Zone Matrix (GLSZM)} (16 features), \textit{Neighbouring Gray Tone Difference Matrix} (5 features), and \textit{Gray Level Dependence Matrix (GLDM)} (14 features). To capture localized tissue characteristics, we apply spatial decomposition by dividing each pathological ROI into patches at multiple granularities: a single patch (\(1 \times 1 \times 1\), i.e., using the entire ROI as input), 8 patches (\(2 \times 2 \times 2\)), and 27 patches (\(3 \times 3 \times 3\)). For each configuration, features are independently extracted across axial, coronal, and sagittal sequences, resulting in a high-dimensional radiomic feature pool that comprehensively represents 3D tissue characteristics. A 3DResNet-18~\cite{he2016deep} serves as the feature-weighting neural network, learning patient-specific relevance scores to adaptively select discriminative features for individual cases.

\noindent\textbf{Classification Model:}  
As described in Sec.~\ref{sec:method}, the weighted feature vector $\mathbf{f}^{\text{w}}$ is passed to a classifier \(R_{\psi}\) to predict class probabilities. In this work, we implement \(R_{\psi}\) as an extended logistic regression model that incorporates both first-order individual feature contributions and second-order pairwise interactions. This design enables the classifier to capture second-order dependencies among selected features, should they be predictive for the classification, while maintaining relatively high explainability. During model development, values of the task-specific classification thresholds $T$ are determined based on the best Youden’s index obtainable on the validation set. 

\begin{table}[!ht]
\renewcommand{\thetable}{1}
\centering
\caption{Comparison with end-to-end DL models. N: Number of patches; PFS: Patient-specific features selection; Reg: Registration in preprocessing. 
* denotes reproduced results}
\label{tab1}

\renewcommand{\arraystretch}{0.9} 

\setlength{\tabcolsep}{1.3pt}
\fontsize{8pt}{9pt}\selectfont

\begin{tabular}{|@{\hskip 0.9pt}c|@{\hskip 0.9pt}c|@{\hskip 0.9pt}c|@{\hskip 0.9pt}c|@{\hskip 0.9pt}c|@{\extracolsep{1.5pt}}c|c@{\hskip 1.5pt}c@{\hskip 1.5pt}c@{\hskip 1.5pt}c|}

\hline
 & & & & & & \multicolumn{4}{c|}{\textbf{Evaluation Metrics}} \\ 
\cline{7-10} 
\textbf{Method} & \textbf{N} & \textbf{PFS}  & \textbf{View} & \textbf{Reg} & \textbf{Type} & \textbf{Acc} & \textbf{Sen} & \textbf{Spe} & \textbf{AUC} \\ 
\hline

\hline
\textbf{\multirow{3}{*}{Ours}} &\multirow{3}{*}{2*2*2} & \multirow{3}{*}{\ding{51}} &  \multirow{3}{*}{3} & \multirow{3}{*}{\ding{51}} 
& abn & \textbf{0.92±0.09} & 0.98±0.03 & 0.71±0.33 & 0.85±0.17 \\
&&&&& acl & \textbf{0.94±0.09} & 0.96±0.06 & 0.88±0.05 & 0.92±0.04 \\
&&&&& men & \textbf{0.84±0.16} & 0.97±0.04 & 0.67±0.34 & 0.82±0.18 \\
\hline


\multirow{3}{*}{\textbf{MRNet*}} & \multirow{3}{*}{-}  & \multirow{3}{*}{-} & \multirow{3}{*}{3} & \multirow{3}{*}{-} 
& abn & 0.85±0.01 & 0.84±0.03 & 0.65±0.09 & 0.94±0.02 \\
&&&&& acl & 0.86±0.02 & 0.77±0.02 & 0.97±0.02 & 0.97±0.02 \\
&&&&& men & 0.71±0.04 & 0.69±0.03 & 0.74±0.03 & 0.84±0.04 \\

\hline
\multirow{3}{*}{\textbf{ELNet*}} & \multirow{3}{*}{-} & \multirow{3}{*}{-} & \multirow{3}{*}{3} & \multirow{3}{*}{-} 
& abn & 0.80±0.01 & 0.95±0.01 & 0.22±0.03 & 0.73±0.01 \\
&&&&& acl & 0.70±0.10 & 0.72±0.26 & 0.68±0.04 & 0.73±0.08 \\
&&&&& men & 0.65±0.06 & 0.61±0.13 & 0.63±0.07 & 0.69±0.03 \\
\hline
\multirow{3}{*}{\textbf{SKID}} & \multirow{3}{*}{-}  & \multirow{3}{*}{-} & \multirow{3}{*}{3} & \multirow{3}{*}{-} 
& abn & 0.87 & 0.98 & 0.49 & 0.90 \\
&&&&& acl & 0.80 & 0.74 & 0.85 & 0.89 \\
&&&&& men & 0.73 & 0.92 & 0.57 & 0.81 \\
\hline

\end{tabular}
\end{table}

\begin{table}[!ht]
\renewcommand{\thetable}{2}
\centering
\caption{Ablation study evaluating different configuration settings. N: Number of patches; RF: Radiomics features PFS: Patient-specific features selection; Reg: Registration in preprocessing. 
}
\label{tab1}

\renewcommand{\arraystretch}{0.9} 

\setlength{\tabcolsep}{1.3pt}
\fontsize{8pt}{9pt}\selectfont

\begin{tabular}{|@{\hskip 0.9pt}c|@{\hskip 0.9pt}c|@{\hskip 0.9pt}c|@{\hskip 0.9pt}c|@{\hskip 0.9pt}c|@{\hskip 0.9pt}c|@{\extracolsep{1.5pt}}c|c@{\hskip 1.5pt}c@{\hskip 1.5pt}c@{\hskip 1.5pt}c|}

\hline
 & & & & & & & \multicolumn{4}{c|}{\textbf{Evaluation Metrics}} \\ 
\cline{8-11} 
\textbf{Method} & \textbf{N} & \textbf{RF Type} & \textbf{FS} & \textbf{View} & \textbf{Reg} & \textbf{Type} & \textbf{Acc} & \textbf{Sen} & \textbf{Spe} & \textbf{AUC} \\ 
\hline

\hline
\textbf{\multirow{3}{*}{Ours}} &\multirow{3}{*}{2*2*2} & \multirow{3}{*}{All} & \multirow{3}{*}{\ding{51}} & \multirow{3}{*}{3} & \multirow{3}{*}{\ding{51}} 
& abn & \textbf{0.92±0.09} & 0.98±0.03 & 0.71±0.33 & 0.85±0.17 \\
&&&&&& acl & \textbf{0.94±0.09} & 0.96±0.06 & 0.88±0.05 & 0.92±0.04 \\
&&&&&& men & 0.84±0.16 & 0.97±0.04 & 0.67±0.34 & 0.82±0.18 \\
\hline

\textbf{\multirow{3}{*}{Ours}} & \multirow{3}{*}{3*3*3} & \multirow{3}{*}{All} & \multirow{3}{*}{\ding{51}} & \multirow{3}{*}{3} & \multirow{3}{*}{\ding{51}} 
& abn & 0.91±0.11 & 0.98±0.08 & 0.65±0.41 & 0.83±0.21 \\
&&&&&& acl & 0.94±0.11 & 0.89±0.20 & 0.98±0.04 & 0.94±0.12 \\
&&&&&& men & 0.83±0.16 & 0.74±0.31 & 0.89±0.22 & 0.81±0.16 \\

\hline
\multirow{3}{*}{2nd RF} & \multirow{3}{*}{2*2*2} & \multirow{3}{*}{2nd} & \multirow{3}{*}{\ding{51}} & \multirow{3}{*}{3} & \multirow{3}{*}{\ding{51}} 
& abn & 0.90±0.08 & 0.98±0.03 & 0.68±0.25 & 0.83±0.14 \\
&&&&&& acl & 0.83±0.17 & 0.91±0.12 & 0.74±0.26 & 0.82±0.18 \\
&&&&&& men & \textbf{0.88±0.14} & 0.92±0.13 & 0.85±0.16 & 0.88±0.13 \\
\hline
\multirow{3}{*}{1st RF} & \multirow{3}{*}{2*2*2} & \multirow{3}{*}{1st+shape} & \multirow{3}{*}{\ding{51}} & \multirow{3}{*}{3} & \multirow{3}{*}{\ding{51}} 
& abn & 0.77±0.13 & 0.75±0.19 & 0.87±0.18 & 0.81±0.10 \\
&&&&&& acl & 0.82±0.14 & 0.79±0.19 & 0.83±0.13 & 0.81±0.14 \\
&&&&&& men & 0.72±0.10 & 0.80±0.15 & 0.68±0.09 & 0.74±0.10 \\
\hline
\multirow{3}{*}{1patch}& \multirow{3}{*}{1*1*1} & \multirow{3}{*}{All} & \multirow{3}{*}{\ding{51}} & \multirow{3}{*}{3} & \multirow{3}{*}{\ding{51}} 
& abn & 0.73±0.32 & 0.66±0.44 & 0.93±0.13 & 0.79±0.21 \\
&&&&&& acl & 0.76±0.14 & 0.97±0.05 & 0.58±0.24 & 0.77±0.12 \\
&&&&&& men & 0.62±0.07 & 0.78±0.25 & 0.55±0.27 & 0.66±0.04 \\
\hline

\multirow{3}{*}{NoFS}& \multirow{3}{*}{2*2*2} & \multirow{3}{*}{All} & \multirow{3}{*}{\ding{55}} & \multirow{3}{*}{3} & \multirow{3}{*}{\ding{51}} 
& abn & 0.88±0.12 & 0.89±0.14 & 0.90±0.22 & 0.89±0.13 \\
&&&&&& acl & 0.92±0.04 & 0.96±0.05 & 0.93±0.12 & 0.94±0.08 \\
&&&&&& men & 0.83±0.17 & 0.90±0.07 & 0.81±0.24 & 0.85±0.14 \\
\hline

\multirow{3}{*}{NoReg}& \multirow{3}{*}{2*2*2} & \multirow{3}{*}{All} & \multirow{3}{*}{\ding{51}} & \multirow{3}{*}{3} & \multirow{3}{*}{\ding{55}} 
& abn & 0.77±0.08 & 0.59±0.13 & 0.95±0.04 & 0.75±0.07 \\
&&&&&& acl & 0.71±0.09 & 0.43±0.18 & 0.98±0.04 & 0.68±0.10 \\
&&&&&& men & 0.65±0.14 & 0.97±0.04 & 0.43±0.26 & 0.69±0.13 \\
\hline

\end{tabular}
\end{table}

\noindent\textbf{Comparison and Ablation Experiments:}  
We compared our method with established end-to-end DL models, including MRNet~\cite{bien2018deep}, ELNet~\cite{tsai2020knee}, and SKID~\cite{manna2023self}. To analyze the impact of core components, we conducted ablation studies focusing on three aspects: radiomic feature composition, feature selection mechanism, and spatial decomposition. Using the full radiomic feature set consistently outperformed configurations restricted to second-order or first-order \& shape features. Disabling the feature-weighting network and using unweighted features reduced accuracy, confirming the value of adaptive selection. Finally, among various patch sizes, the \(2 \times 2 \times 2\) configuration delivered the most favorable balance between diagnostic performance and model explainability (specific examples are discussed in Sec.~\ref{sec:results}).

\begin{figure}
    \centering
     \includegraphics[width=1\textwidth]{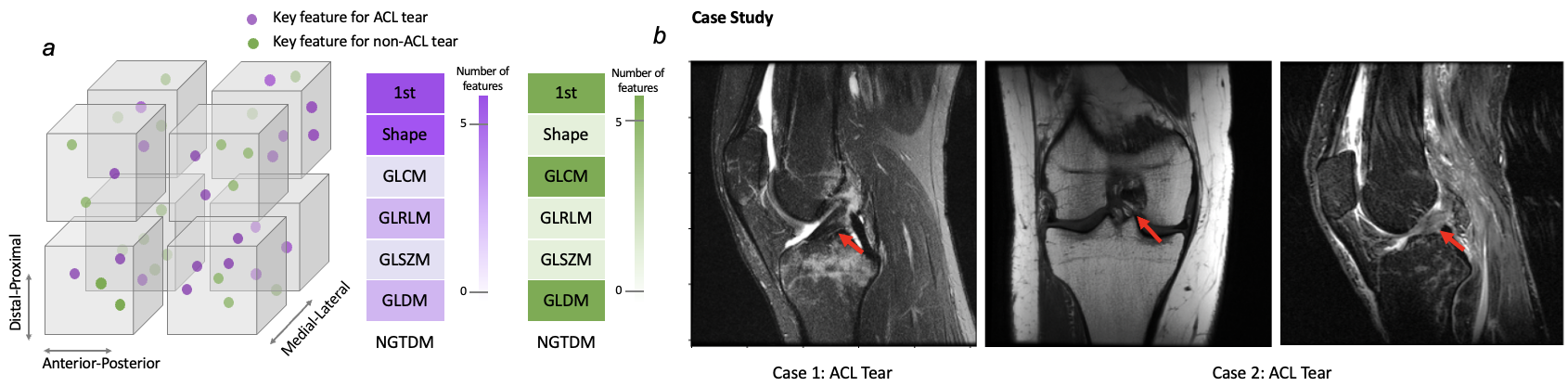}
    \caption{(a) Anatomical location and feature type distribution of the top 20 top radiomic features; (b) representative cases of ACL tears (details in Sec.\ref{sec:results}).}
    \label{fig2}
\end{figure}

\subsection{Results and Interpretation}
\label{sec:results}

\noindent\textbf{Quantitative Evaluation:}
For ACL tear classification, our patient-specific radiomics signature framework achieves an accuracy of \(0.94 \pm 0.09\), AUC of \(0.92 \pm 0.04\), and sensitivity of \(0.96 \pm 0.06\). Compared to MRNet and ELNet, our method demonstrates consistently strong performance. Notably, statistically significant improvements were observed in specificity (vs. ELNet, \(p = 0.0002\)) and in accuracy (vs. ELNet, \(p = 0.012\)). For meniscus tear classification, our method also achieves superior performance, particularly in accuracy (\(p = 0.028\) vs. ELNet), highlighting its robustness across multiple diagnostic tasks. 

Table~\ref{tab1} also presents key ablation results. When evaluating radiomic feature types, the full feature set (first-order, shape, and texture features) consistently yields the highest performance. Limiting the model to only second-order features slightly degrades overall accuracy, while using only first-order and shape features results in notable performance drop. Removing the feature-weighting neural network and using unweighted features (NoFS) leads to a noticeable decline in performance, confirming the critical role of adaptive feature selection. 

\noindent\textbf{Feature Analysis \& Case Study:} Based on feature analysis of the ACL tear test set, the most relevant radiomic features primarily originate from the sagittal view, which aligns with clinical practice, where ACL tears are most clearly visualized. When predicting the presence of an ACL tear, key features tend to concentrate in patches that correspond anatomically to the ACL location. However, in some cases, key features appear dispersed, reflecting secondary structural changes — such as bone marrow edema and anterior tibial translation — that extend beyond the ligament itself and affect the surrounding tissues. In cases of ACL tear, the top-ranked features are predominantly first-order statistics and shape-based features, with additional contribution from texture features. These features effectively capture macrostructural abnormalities, including ligament discontinuity, joint effusion, and altered signal intensity. In contrast, non-ACL tear cases appear to rely more on texture features, particularly \textit{GLCM} and \textit{GLDM}. This suggests that, when there is no obvious structural disruption, texture analysis may provide valuable information about subtle tissue heterogeneity.

We further analyse two representative cases with differing severities of ACL injury. Case 1 shows residual fibres attached to the tibial insertion, while Case 2 demonstrates loss of normal fibre architecture (as indicated by red arrows), accompanied by more extensive bone marrow edema and joint effusion. In both cases, the most important radiomic features include Large Area Low Gray Level Emphasis, Gray Level Variance, and Sum Entropy. These features reflect, respectively, the extent of low-intensity regions associated with joint fluid, variability in signal due to disrupted fibre structure, and increased complexity of signal distribution within the lesion zone. The most notable feature between the two cases is Large Dependence Low Gray Level Emphasis (LDLGLE), which quantifies large, homogeneous low-intensity areas. The case with more widespread disruption shows higher LDLGLE values, consistent with a greater extent of uniform signal alteration. This distinction approximates radiological interpretation, which is influenced by differences in the extent of ligamentous injury.



\section{Conclusion \& Discussion}

We propose a deep learning–guided, patient-specific radiomics framework for knee MRI diagnosis that integrates classical handcrafted features with modern data-driven modeling. By dynamically selecting relevant features tailored to individual imaging characteristics and applying neural weighting, the method leverages the interpretability of traditional radiomics while benefiting from the representational power of deep learning. The final transparent logistic regression classifier ensures interpretability, offering clinicians actionable insights into which features drive each prediction—an essential quality often lacking in purely deep learning–based medical imaging models.

While our framework improves interpretability, radiomic features may still be sensitive to variations in imaging protocols, such as scanner type and sequence settings. These differences can introduce non-biological variability, affecting generalizability. To address this, we incorporated image registration to reduce anatomical differences across scans. Nevertheless, additional harmonization strategies—such as domain adaptation—may be necessary to ensure further robustness across diverse imaging environments.

\section{Acknowledgement}
This work was supported by the International Alliance for Cancer Early Detection, an alliance between Cancer Research UK [EDDAPA-2024/100014] \& [C73666/A31378], Canary Center at Stanford University, the University of Cambridge, OHSU Knight Cancer Institute, University College London and the University of Manchester; and the National Institute for Health Research University College London Hospitals Biomedical Research Centre.

%
%
%
%

\bibliographystyle{splncs04}
\bibliography{MICCAI2025_paper_template}

\end{document}